\documentclass{article}

\usepackage{natbib}

\usepackage[preprint]{neurips_2025}

\usepackage[utf8]{inputenc} 
\usepackage[T1]{fontenc}    
\usepackage{hyperref}       
\usepackage{url}            
\usepackage{booktabs}       
\usepackage{amsfonts}       
\usepackage{nicefrac}       
\usepackage{microtype}      
\usepackage{xcolor}         
\usepackage{amsmath}
\usepackage{graphicx}
\usepackage[linesnumbered, ruled]{algorithm2e}
\usepackage{float}
\usepackage{tabularx}
\usepackage{subcaption} 
\usepackage{amsthm}
\newtheorem{problem}{Problem}[section] 
\usepackage[most]{tcolorbox}

\title{Automatic Demonstration Selection for LLM-based Tabular Data Classification}

\author{%
  Shuchu Han, Wolfgang Bruckner\\
  Capital One\\
  \texttt{\{shuchu.han, wolfgang.bruckner\}@capitalone.com  } \\
}

\begin{document}

\maketitle

\begin{abstract}
A fundamental question in applying In-Context Learning (ICL) for tabular data classification is how to determine the ideal number of demonstrations in the prompt. This work addresses this challenge by presenting an algorithm to automatically select a reasonable number of required demonstrations. Our method distinguishes itself by integrating not only the tabular data's distribution but also the user's selected prompt template and the specific Large Language Model (LLM) into its estimation. Rooted in Spectral Graph Theory, our proposed algorithm defines a novel metric to quantify the similarities between different demonstrations. We then construct a similarity graph and analyze the eigenvalues of its Laplacian to derive the minimum number of demonstrations capable of representing the data within the LLM's intrinsic representation space. We validate the efficacy of our approach through experiments comparing its performance against conventional random selection algorithms on diverse datasets and LLMs.
\end{abstract}

\section{Introduction}
Within enterprise settings, tabular data is the most prevalent form of information. The emergence of Generative AI has prompted researchers and organizations to investigate methods for leveraging Large Language Models (LLMs) to derive insights from data stored in Relational Database Management Systems (RDBMS)~\cite{codd1970relational}. Initial efforts in this domain involved the fine-tuning of pre-trained LLMs for domain-specific learning objectives~\cite{dinh2022lift}~\cite{hu2022lora}~\cite{houlsby2019parameter}~\cite{rebuffi2017learning}~\cite{wang2020k}. Subsequently, the research community proposed In-Context Learning (ICL)~\cite{brown2020language}~\cite{nam2023semi}~\cite{hegselmann2023tabllm}~\cite{lu2021fantastically}~\cite{min2023recent}, a paradigm wherein an LLM acquires new task capabilities solely through the processing of examples and instructions embedded within the input prompt (i.e., the "context"), thereby obviating the need for modifications to the LLM's internal weights or neural network architecture. This inherent simplicity and cost-efficiency render ICL a highly accessible approach for executing classical machine learning tasks, such as classification and prediction, utilizing LLMs, particularly for users without prior specialized machine learning training.

The core philosophy of applying LLMs to tabular data involves representing the table in a natural language format, effectively "teaching" the LLM to interpret it as a human would. This mimicry has led to observed improvements in classification and prediction tasks on tabular data~\cite{borisov2022language}~\cite{hegselmann2023tabllm}~\cite{yan2024making}. However, a fundamental question remains for enhancing user experience with ICL: How many demonstrations should be included in the prompt? The common practice, as noted in many papers, is to randomly select a small number of samples (e.g., two, three, or five) to construct these demonstrations. While seemingly straightforward, this question presents several challenges. For instance, the specific learning task (classification or prediction) may be unknown, and the nature of the target variable (discrete category or continuous numerical value) varies. Furthermore, considerations such as data quality, missing values, and the possibility of partial columns in end-user input when utilizing LLM services add further complexity.

In this work, we try to tackle a small portion of above mentioned questions. To be specific, We formally define our problem as:
\begin{problem}
    Given a pre-trained LLM and a tabular dataset, can we select few demonstrations automatically for the users with quality classification performance in general. 
\end{problem}
While our ultimate aim is to address a general problem, we prioritize refining it into a \textbf{demonstration selection problem}. There are few challenges behind our defined problem:

\paragraph{The impact of labels} The research work~\cite{min2022rethinking} shows that the golden labels actually do not have significant impact to the classification performance. Unlike the classical machine learning, if the given dataset is separable in the feature vector space, a subset of randomly selected samples from each category can have the better performance than random samples after the  machine learning model is trained, for example, support vector machine (SVM). Nevertheless, for the learning task we are discussing in this work, we don't limit the user to only select one column as the label column. The users have the freedom to choose arbitrary column as the label column. 

\paragraph{Lack of theory support} To our knowledge, there is no well proved theory about how to select good demonstrations from a given sample set for a given LLM. Most existing works~\cite{wang2023large}~\cite{qin2023context}~\cite{zebaze2024context}are treating the LLMs as machine learning models, and employing the well established theory framework like Information theory~\cite{shannon1948mathematical},  Solomonoff Induction~\cite{solomonoff1964formal}, Minimum Description Length~\cite{grunwald2007minimum} or etc. 

With above mentioned problems and challenges, we presented a heuristic algorithm to select demonstrations automatically for the ICL over tabular data. We treat each row of the tabular data as a demonstration candidate, and estimate the data distribution of the candidates in the representation space of the LLM. To be specific, we use the token IDs as the representation of each demonstration, and use a sparse graph to represent the data distribution. With that, we use the Spectral Gap from the Spectral graph theory to estimate the number of clusters and use it as the minimum number of demonstrations. Our contribuions are summarized as follows:
\begin{enumerate}
    \item We proposed a heuristic algorithm to select demonstrations automatically for ICL with tabular data, Especially for classification tasks.
    \item Our algorithm is almost a parametric-free algorithm, except the construction of the sparse graph. 
    \item Our empirical studies show that our proposed algorithm can have resonable classification performance stablely on different LLMs and datasets.
\end{enumerate}

\begin{figure}[ht!]
\centering
\includegraphics[width=\textwidth]{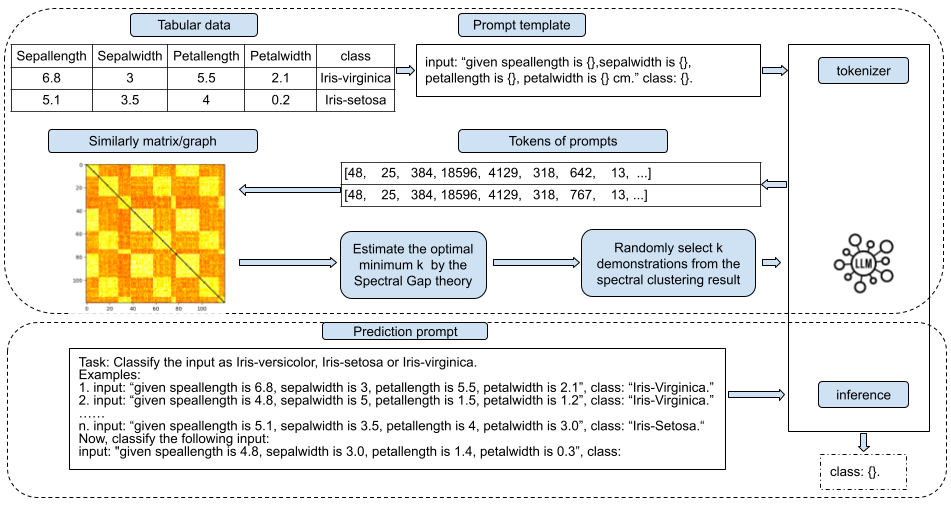}
\caption{An overview of our proposed algorithm. The prompt template and the LLM are chosen by the user.}
\end{figure}

\section{Algorithm}
In this section, we present the details of our proposed algorithm. We first present the demonstration design for each row of the tabular data. Secondly, we calculate the similarity among different demonstrations and build a similarity graph. Thirdly, we use the spectral graph theory to estimate the number of clusters in the similarity graph, which is the number of demonstrations we need. Finally, we design a prompt template with the selected demonstrations for the final inference.

\subsection{Demonstration Representation}
For in-context learning over the tabular data, we need to package the information held by each row of the table into the language that can be understood by the LLM. Without loss of generality, we follow the design from ~\cite{borisov2022language}~\cite{dinh2022lift}, and design our demonstration as follows:
\begin{center}
    "input: given \{column name\} is \{column value\}, $\dots$, and \{column name\} is \{column value\}, class: \{label\}."
\end{center}
An example illustration for the Iris dataset~\cite{iris_53} is: 
\begin{center}
    "input: given \underline{sepallength} is \underline{6.8}, \underline{sepalwidth} is \underline{3.0}, \underline{petallength} is \underline{5.5}, and \underline{petalwidth} is \underline{2.1}, class: \underline{Iris-virginica}."
\end{center}
As we can see from the above example, we don't explicit add modification to the original table data like improve the column names from "sepallength" to "sepal length" which is more human readable and meaningful. Similarity, we don't add extra description to the table values like use "6.8 cm" instead of "6.8". In our proposed algorithm, the end-users always have the freedom to choose a better demonstration template based on their expertise.

\subsection{Similarity Metric}
To build the similarity graph, we need to find a reasonable similarity metric to measure the difference among demonstrations. The most popular choice is to use the embedding vector of the whole string of the demonstration. However, we select the token IDs in our algorithm. The reason is two fold: 1) the embedding vectors of a pre-trained LLM are usually high dimensional vectors. Any general similarity metric such as the cosine similarity will be affected by the Curse of Dimensionality. As a result, almost all calculated similarity values will stay in a small value range and are very difficult to differentiate. Moreover, one demonstration can have many tokens (as the nature of tabular data). By using the aggregated vector to represent the demonstration will make the situation even worse. 2) the usage of high dimensional embedding vectors is computation heavy in terms of CPU/GPU and memory. 

For the goal of efficiency, we use token IDs and the Jaccard Similarity as the similarity metric. Given a pre-trained LLM $\mathcal{M}$, we calculate the similarity of demonstrations as: 
\begin{equation}
    similarity(e_i, e_j) = \frac{\|e_i \cap e_j\|}{\|e_i \cup e_j\|},
\end{equation}
where $e_i$ is the token list of demonstrations $Demo(s_i)$.

\subsection{Similarity Graph}
With the defined similarity metric, we can build our similarity graph $\mathcal{G}$ as:
\begin{equation}
    \mathcal{G}(i, j) = \left\{\begin{array}{rcl}
    similarity(e_i, e_j) & \mbox{for} & i \neq j \\
    0                          & \mbox{for} & i = j \\
    \end{array}\right.
\end{equation}

\subsection{Evaluate the Number of Minimum Demonstrations}
To evaluate the minimum number of demonstrations, we use the spectral gap method~\cite{chung1997spectral} on the Laplacian matrix. Since we are calculating the spectra of a matrix, we hope that the graph can be sparse and can better represent the distribution of the data~\cite{von2007tutorial}. With that, we transfer the similarity graph into a $k$-nearest neighbor graph by only keeping the top-$k$ nodes with highest similarity values. After that, we calculate the Laplacian matrix as follows:
\begin{equation}
    \mathcal{L} =\mathcal{D} - \mathcal{G}
\end{equation}
where $D$ is the degree matrix:
\begin{equation}
    \mathcal{D}_{ii} = \sum_{j}\mathcal{G}_{ij}
\end{equation}

The normalize the graph Laplacian matrix is:
\begin{equation}\label{eq:nm_laplacian}
    \mathcal{L}_{sys} = \mathcal{D}^{-1/2}\mathcal{L}\mathcal{D}^{-1/2}.
\end{equation}

Then we apply the spectral gap method to find the optimal number of clusters, which will be the optimal number of demonstrations in our algorithm.
\begin{equation}\label{eq:spectral_gap}
    k = \arg \max_{\substack{i}} {\lambda_{i+1}-\lambda_{i}}
\end{equation}

The overall algorithm can be summarized as Algorithm~\ref{alg:all}:
\SetKwComment{Comment}{/* }{ */}
\begin{algorithm}[ht!]
\caption{Estimating the minimum number of required prompts}\label{alg:all}
\KwIn{1) tabular data $\mathcal{S}$ with $n$ rows, where each row is a sample $s_i \in \mathcal{S}, i=(1, \dots, n)$\; 2) A demonstration template $Demo()$\; 3) A prediction template; 4) A selected per-trained LLM model: $\mathcal{M}$.} 5) a hyperparameter $k$ for building the sparse similarity graph.\;
\KwOut{The minimum number of required prompts: $d$}
\For{$s_i \in \mathcal{S}$}{Generate the demonstration: $Demo(s_i)$}
\ForEach{$Demo(s_i)$}{Calculate the token ID list $Token(s_i)=[t_0, \cdots, t_j, \cdots]$ by using the tokenizer of the LLM model $\mathcal{M}$}
Build the similarity matrix/graph $\mathcal{G}$ with each node is a token list $Token(s_i)$\;
Creat the sparse similarity graph by using $k$-nearest neighbors, and reset the edge weight to 1.\;
Calculate the normalized Laplacian according to Eq.\eqref{eq:nm_laplacian}\;
Calculate the Eigenvalues: $\lambda_i$\;
Calculate the optimal minimum number of demonstrations $d$ by Eq.\eqref{eq:spectral_gap}\;
Grouping the demonstrations through spectral clustering with number of clusters equal to $d$\;
Select one sample from each group randomly and aggregate them as the minimum set of demonstrations.
\end{algorithm}

\subsection{Demonstration Selection}
To select $d$ demonstrations, we apply the $k$-means algorithm to the eigenvectors corresponding to the smallest $d$ eigenvalues except $\lambda_0$. With the clustering result, we are able to select $d$ demonstration by random sampling one sample from each cluster.

\subsection{Prompt Design}
\begin{tcolorbox}[
    colback=blue!5!white,    
    colframe=blue!75!black,  
    title={\textbf{Prompt template as the input to LLM for the classification inference.}}, 
    arc=3mm,                 
    boxrule=1pt              
    ]
\textbf{Inference Prompt}\\
Task: Classify the input as \{ \}, $\cdots$ or \{ \}. \\
    Examples:
    \begin{enumerate}
        \item "input: given \{ \}, class: \{ \}."
        \item "input: given \{ \}, class: \{ \}."
        \item $\cdots$
    \end{enumerate}
    Now, classify the following input: \\
    input: \{ \}, class:
\label{box:prompt}
\end{tcolorbox}

Under the framework of ICL, we need to design a reasonable prompt for the final inference. There is no golden rule of thumb about how to build the prompt with selected demonstrations. Although different prompt designs may have different inference results as the natural of prompt engineering, we try to follow the rule of simplicity, and use the following prompt~\ref{box:prompt} template to obtain our classification result.

\section{Experiments}
In this section, we present the results of our experiments in validating the correctness and effectiveness of our proposed algorithm. Our experiments can be divided into two parts. The first part is to evaluate the classification performance with different number of randomly selected demonstrations. The second part is to check the performance of the demonstrations selected by our proposed algorithm. All of our experiments are finished on a computer with 12-cores, 64GB memory and a Nvidia 3090ti graphic card. For the sparse graph generation, we build the $k$-nearest neighbor graph and chose $k=10$ as a hyperparameter. The LLM settings, we set the temperature to 0.1 and float value type of Pytorch to float16. 

\subsection{Experiment Design}
We perform our experiments over eight public available tabular datasets from the OpenML website~\cite{van2013openml}. The largest dataset has close to 5k samples (rows). The big data scalability is not our focus here as we always can use down-sampling method to alleviate it. As for the LLM models, we select three different well-known models to examine the performance difference among them. Our major goal is to evaluate the performance difference between the randomly selected demonstrations and the one selected by our proposed algorithm.

\subsection{Datasets}
The details of our selected tabular dataset are described in~\ref{tab:dataset}. Since the content of the prompt is important to the LLM regarding their output, these selected datasets have diverse data formats and column metadata. For example, the dataset \textit{tae} are all Integer values while the \textit{wine} values are all float. Nevertheless, the \textit{Penguins} dataset has mix types which includes numerical values and string values. As for the column metadata, the \textit{iris} dataset has language meaningful column names like "petalwidth" while the \textit{LED} dataset does not have any meaningful column names. We also want to mention that we don't apply any modification to the column names even an easy modification can let the column names are meaningful such as change "petalwidth" to "petal width".

\begin{table}
\centering
\begin{tabularx}{0.9\textwidth}[h]{l|c|c|c|c|c|c|c}
\toprule
Dataset  & \#Sample & \#Feature & \#Category  & (M) & llama & Mistral & Qwen \\
\hline
tae         & 151 & 5 & 3  & Yes & 1 & 4 & 4\\  
cmc         & 1473 & 9 & 3  & Yes& 45 & 42 & 42\\ 
wine        & 178 & 13 & 3 & Yes & 1 & 5 & 5 \\ 
iris        & 150 & 4 & 3  & Yes & 3 & 3 & 3\\ 
penguins & 344 & 6 & 3  & Yes & 5 & 5 & 5 \\  
LED         & 500 & 7 & 10  & No & 5 & 4 & 4\\  
Customers & 440 & 8 & 2  & No & 1 & 3 & 3\\  
spambase  & 4601 & 57 & 2  & Yes & 3 & 1 & 1\\ 

\bottomrule
\end{tabularx}
\caption{Experiment datasets and their properties, where (M) means whether the column names are meaningful or not. The minimum number of demonstrations estimated by our algorithm are listed at the three rightmost columns. There are different numbers for each different LLM.}
\label{tab:dataset}
\end{table}

\subsection{LLM Models}
In our experiment, we select three LLMs form the Huggingface~\cite{huggingface} website with moderate number of parameters and context length by considering the available GPU resource we have.  Context length is the main consideration for this constraint, we required that the LLM can handle at least 10 demonstrations with less than 100 columns. We list the selected LLMs details in Table~\ref{tab:llm_model}
\begin{table}[h!]
\centering
    \begin{tabularx}{0.8\textwidth}{l|c|c|c}
    \toprule
         Model Name& llama & Mistral & Qwen \\
         \hline
         Full name & Llama-3.2-3B & Mistral-7B-v0.3 & Qwen3-8B \\
         \hline
         Context length & 128k & 131k & 8k \\
    \bottomrule
    \end{tabularx}
    \caption{Selected pre-trained LLM models for evaluation.}
    \label{tab:llm_model}
\end{table} 

\subsection{Performance Metric}
Similarly to the work~\cite{min2022rethinking}, we use the Macro-F1 score to evaluate the classification performance since we want to apply our proposed algorithm into general classification tasks. 
\begin{equation}
\text{Macro-}F_1 = \frac{1}{N} \sum_{i=1}^{N} {F_1}_i
\end{equation}
where \( N \) is the number of classes, and \( {F_1}_i \) is the $F_1$ score for the \( i \)-th class.

\subsection{Results}
In this section, we present our experimental results. Our results include two parts. The first part is the evaluation of the classification performance by using randomly selected demonstrations. The second part is the performance comparison between the demonstrations selected by our proposed algorithm and the same number, but randomly selected demonstrations. 

Without loss of generality, we split each dataset into training data and testing data with a potion $80\%$ and $20\%$ respectively. All samples are shuffled before the split. After that, we select demonstrations by using the training data and test the classification performance on all testing data.

\begin{figure}[h!]
    \centering
    \begin{subfigure}[b]{0.24\textwidth}
        \includegraphics[width=\textwidth]{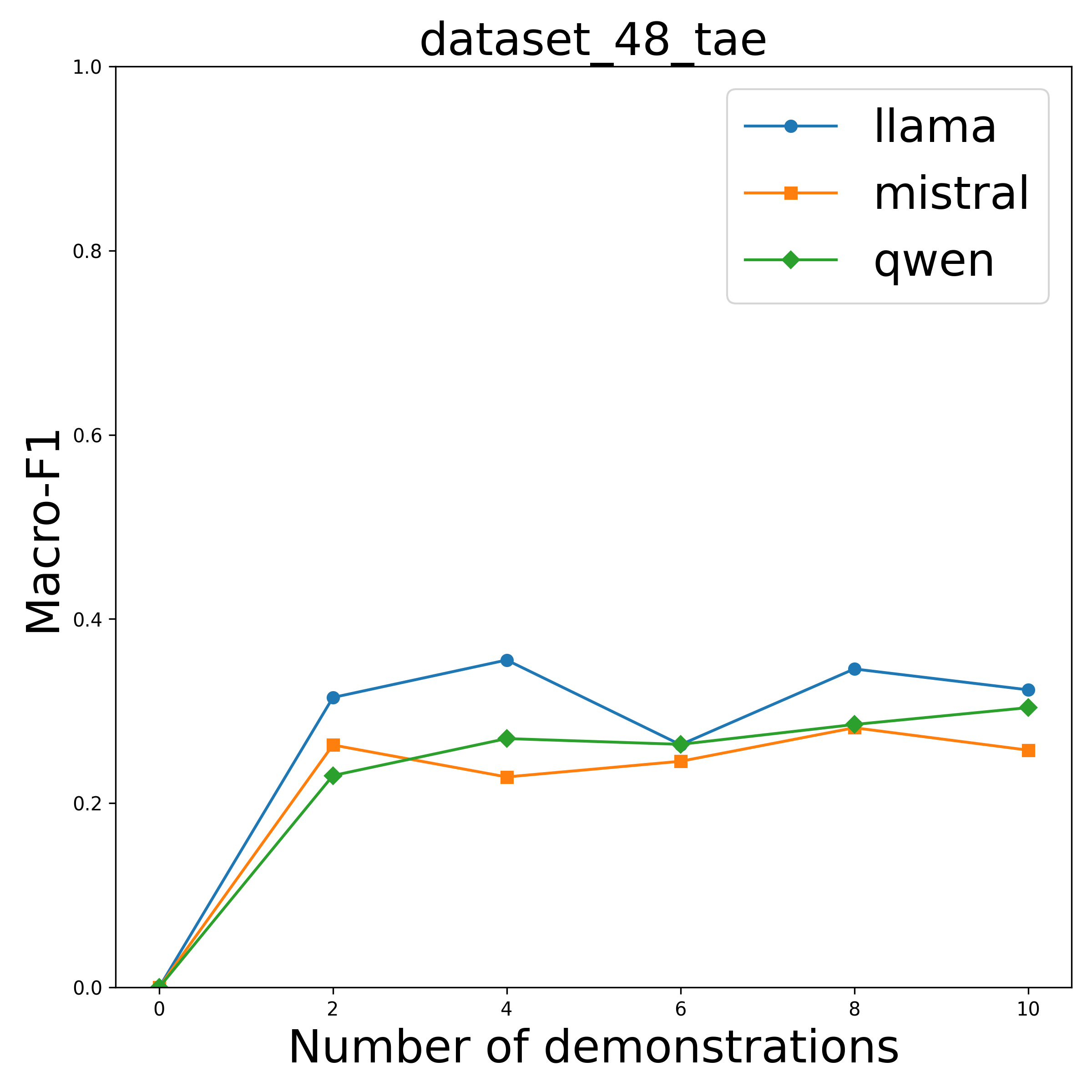}
    \end{subfigure}
    \begin{subfigure}[b]{0.24\textwidth}
        \includegraphics[width=\textwidth]{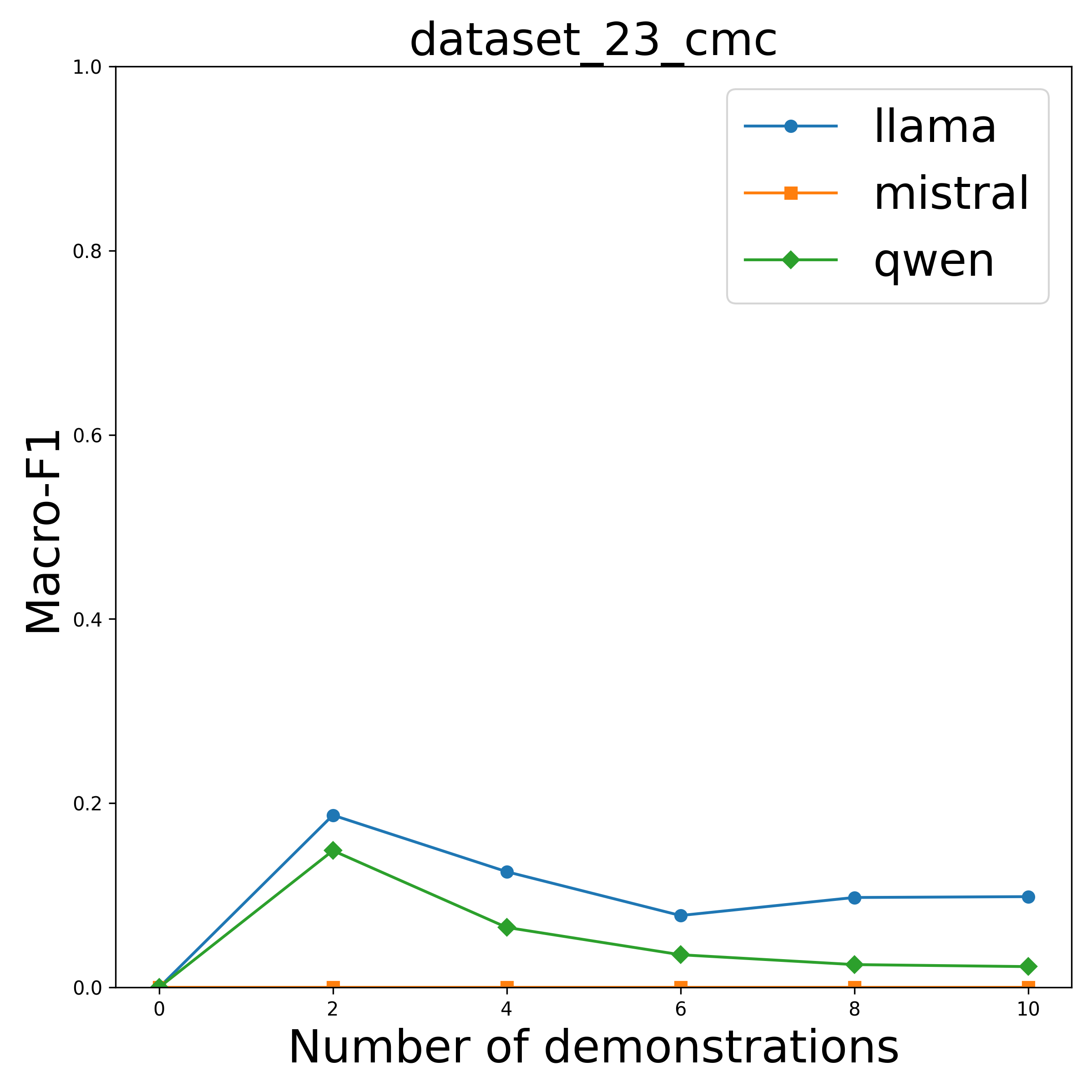}
    \end{subfigure}
        \begin{subfigure}[b]{0.24\textwidth}
        \includegraphics[width=\textwidth]{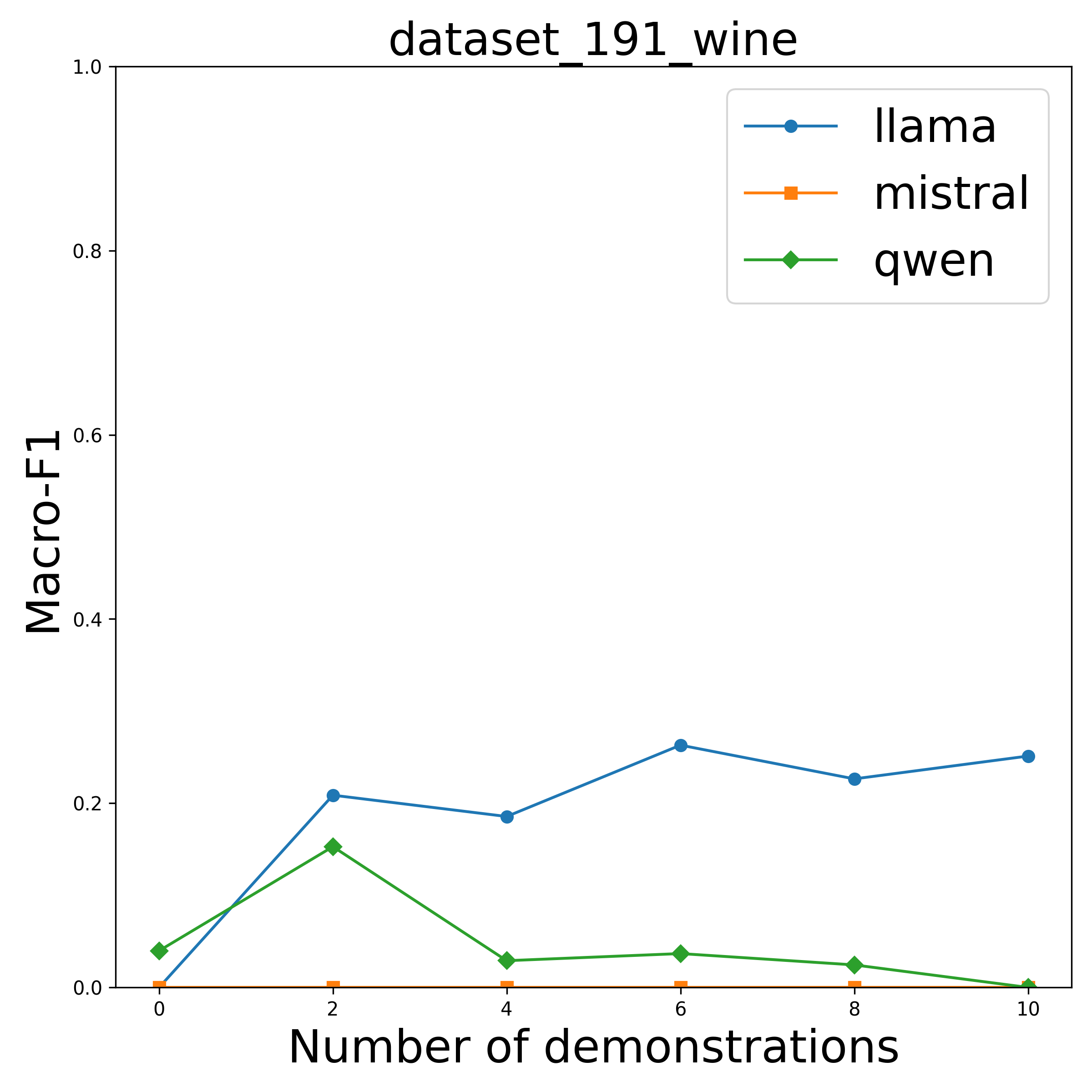}
    \end{subfigure}
        \begin{subfigure}[b]{0.24\textwidth}
        \includegraphics[width=\textwidth]{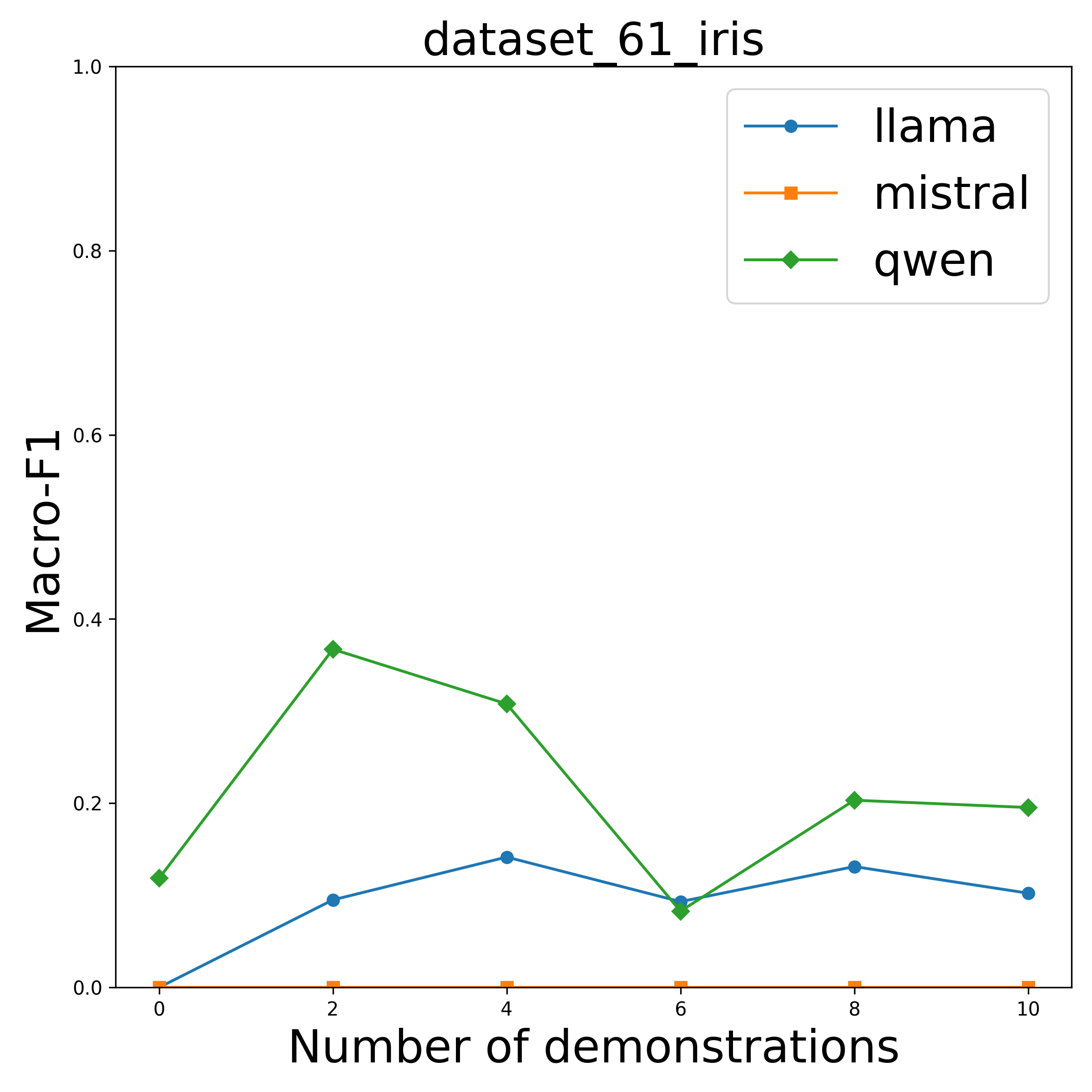}
    \end{subfigure}
    
    \vspace{0.1in} 
    
    \begin{subfigure}[b]{0.24\textwidth}
        \includegraphics[width=\textwidth]{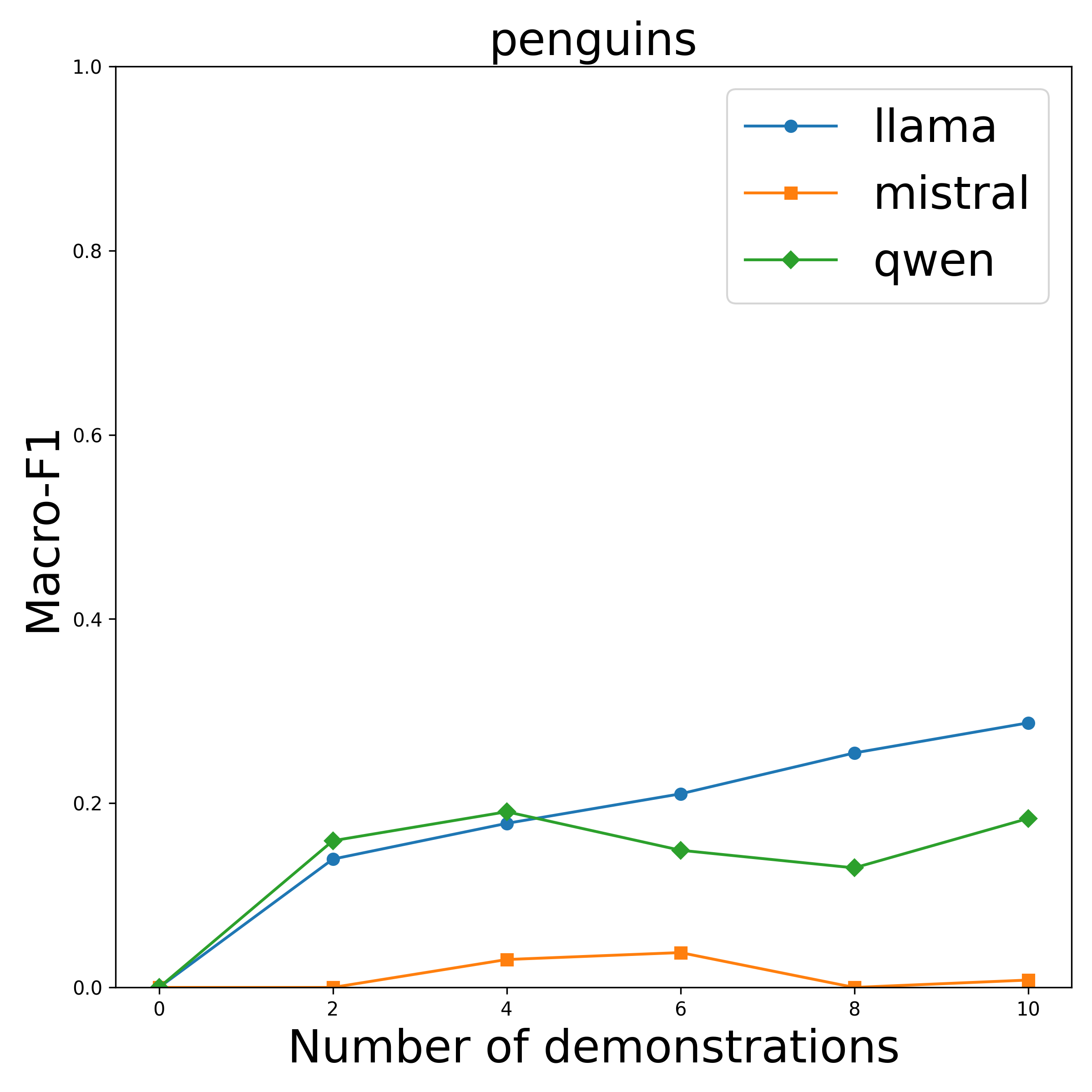}
    \end{subfigure}
    \hfill
    \begin{subfigure}[b]{0.24\textwidth}
        \includegraphics[width=\textwidth]{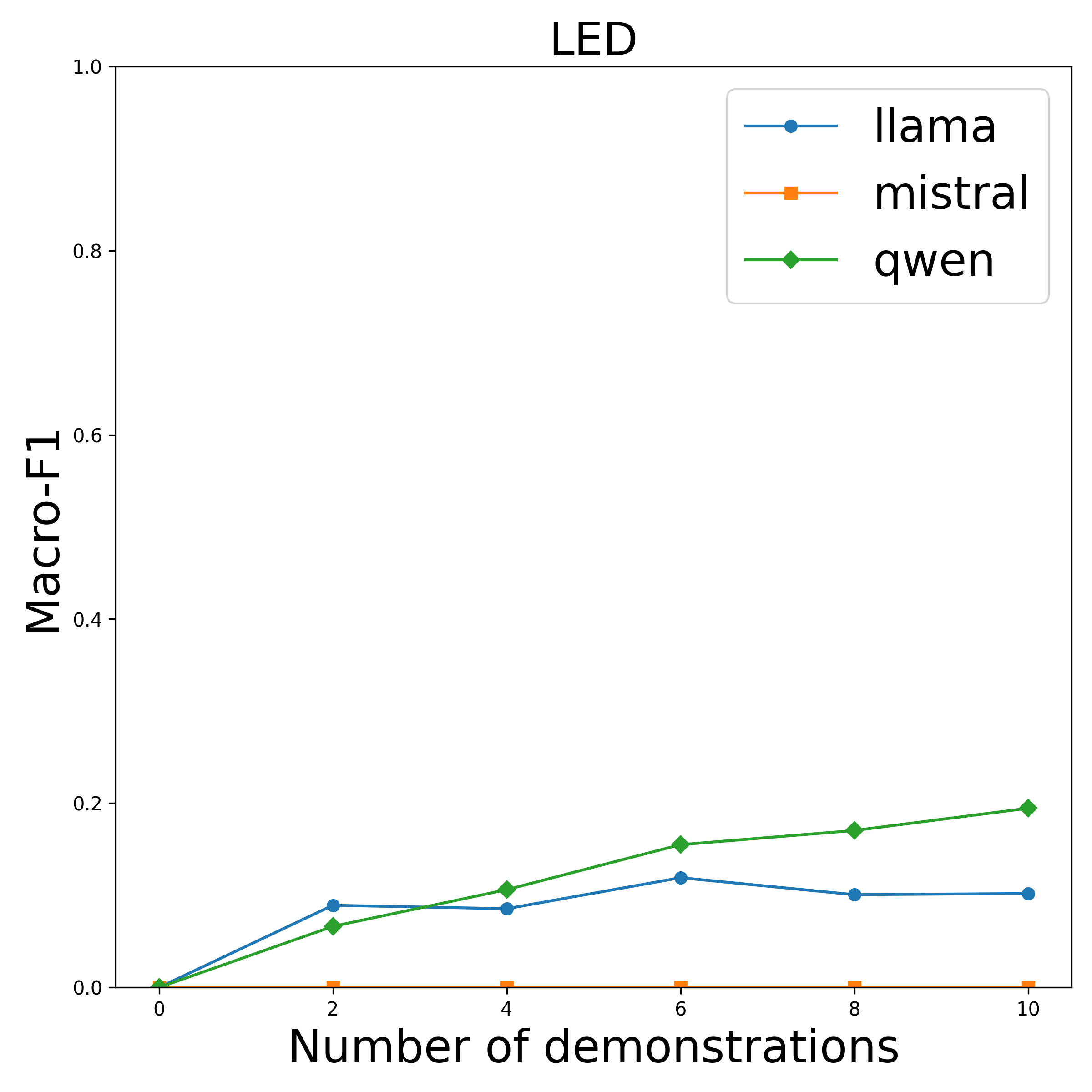}
    \end{subfigure}
    \hfill
        \begin{subfigure}[b]{0.24\textwidth}
        \includegraphics[width=\textwidth]{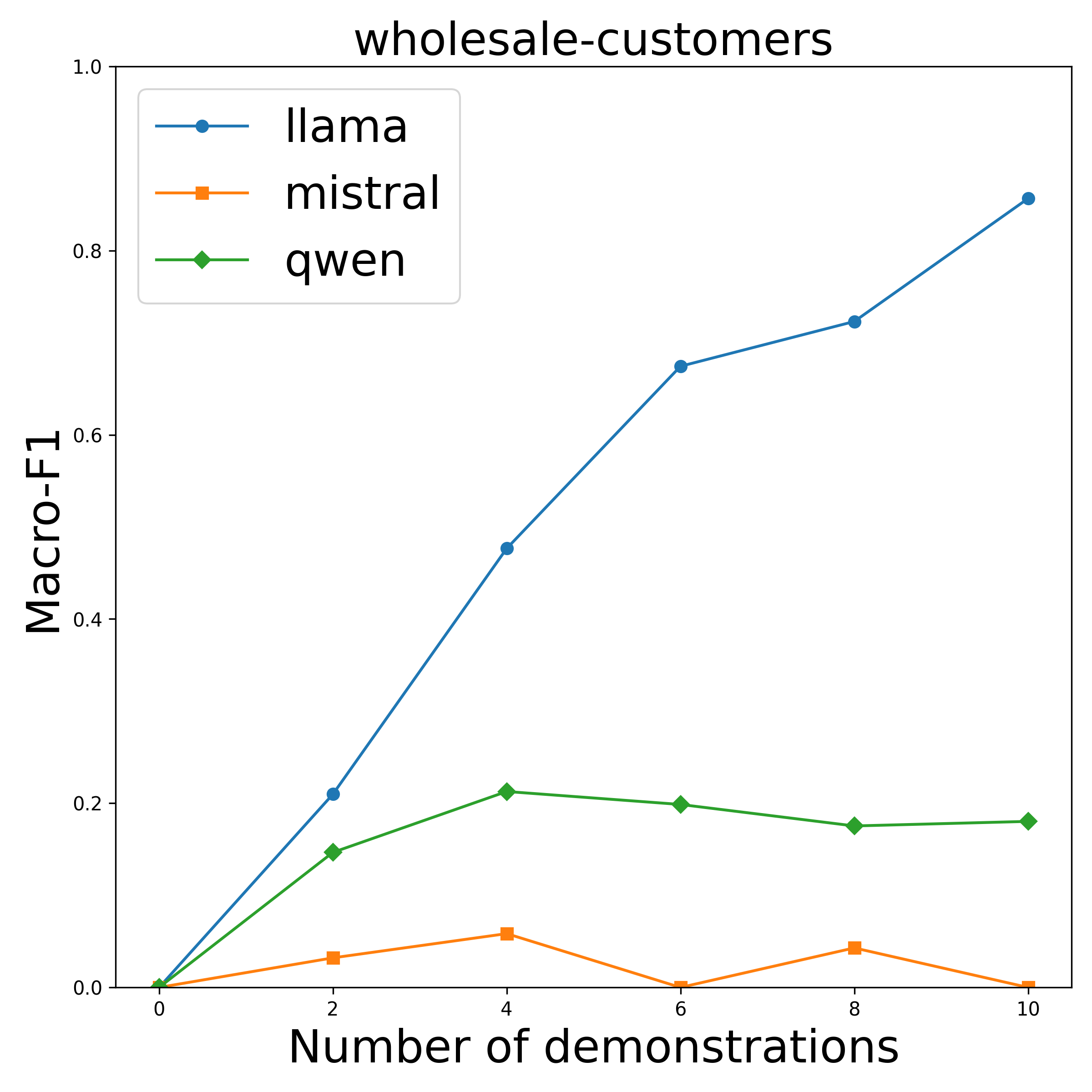}
    \end{subfigure}
    \hfill
        \begin{subfigure}[b]{0.24\textwidth}
        \includegraphics[width=\textwidth]{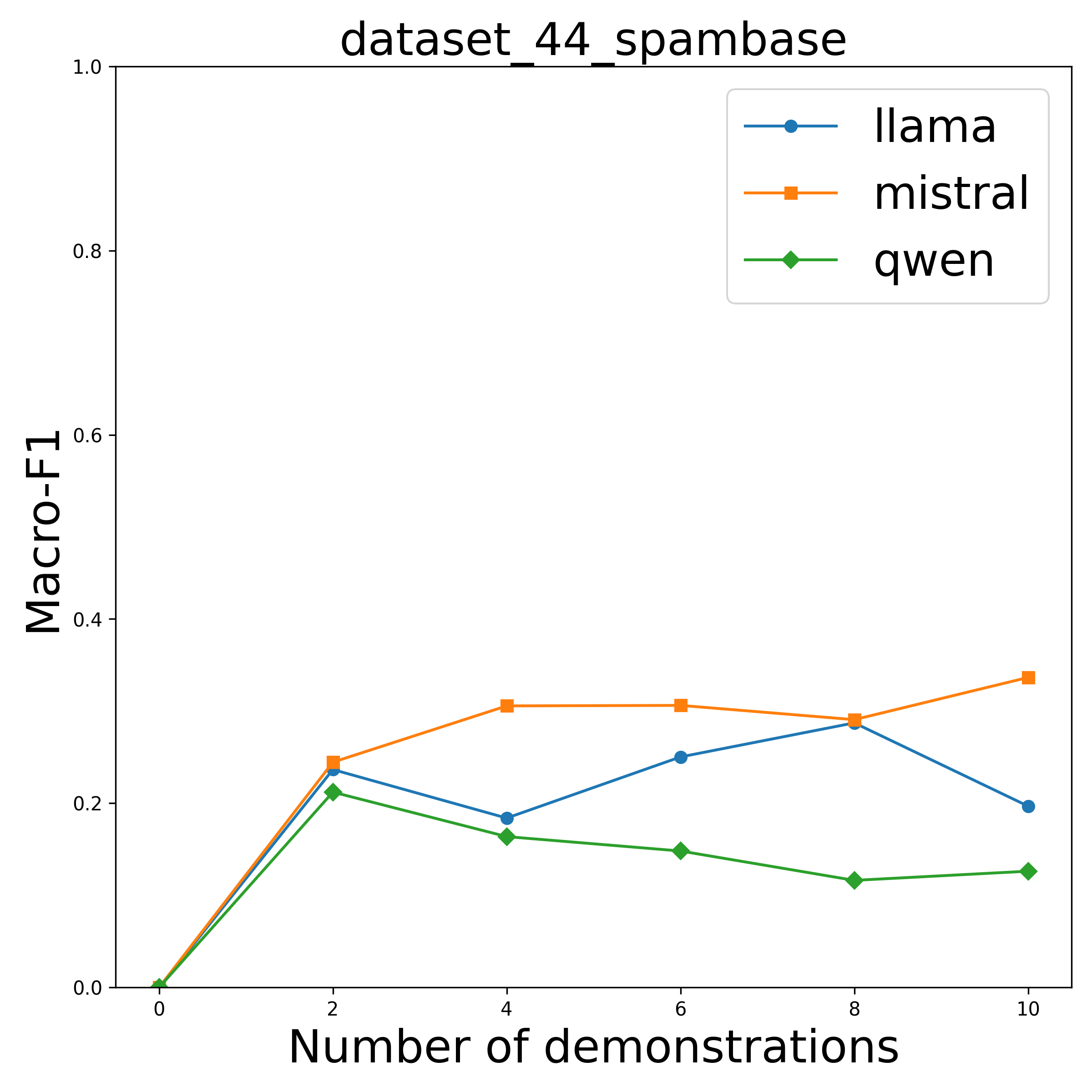}
    \end{subfigure}
    \caption{Classification performance of different number of demonstrations that selected randomly.}
    \label{fig:llm_perf}
\end{figure}

\paragraph{Performance of random selection strategy.} In order to understand the classification performance of randomly selected demonstrations, we evaluate the classification results by selecting different number of demonstrations starting from zero which equals to the zero-shot ICL. To be specific, we randomly select samples from the candidate dataset with different lengths as $[0, 2, 4, 6, 8, 10]$. And we repeat inference on the testing data five times to obtain the average Macro-$F_1$ score. The performance of all datasets are visualized in~\ref{fig:llm_perf}.

From the visualization results, we can see that the classification performance does not necessary increase with more demonstrations. This is confirmed by the observation from other published research works. The only exception is the \textit{wholesale-customers} dataset with the \textit{Llama} model, which is interesting but we can not give a reasonable explanation here unfortunately. 

Another observation is the zero-shot performance. Only the \textit{Qwen} model shows above zero Macro-$F_1$ score on the \textit{Wine} and \textit{iris} datasets. And all other LLMs and datasets have zero scores. Since all the datasets used in our experiment are public available, we guess all these datasets are used as an input to the pre-train process of the LLMs. Based on the results in~\ref{fig:llm_perf}, we can not have a strong argument saying that zero-shot does not work for tabular data classification with LLM. Instead, we can say that few demonstrations do can have better classification performance.

By comparing the results of different LLMs, we can see that the \textit{Mistral} model shows almost failed classification results on dataset \textit{cmc}, \textit{wine}, \textit{iris} and \textit{LED}, while it has much more reasonable performance on the \textit{tae} and \textit{spambase} datasets. The \textit{Llama} and \textit{Qwen} experiments show different performance across different datasets, but give us an impression of overall better performance than the \textit{Mistral} model. With this observation, we can argue that the model selection will also affect the classification performance of tabular data. 

\paragraph{Performance of our algorithm} The rightmost three columns of~\ref{tab:dataset} show the minimum number of demonstrations estimated by our proposed algorithm. We can see that our proposed algorithm shows almost consistent trend among different LLMs, for example, the highest number appears on the \textit{cmc} dataset, the second-highest one is the \textit{tae} dataset and \textit{iris}, \textit{penguins}, \textit{LED} are almost the same. The distribution of the eigenvalues for different datasets is illustrated in Figure~\ref{fig:spectral_gap}. 

Regarding the classification performance, our demonstration selection strategy does not show dominant performance when comparing against random selection. Instead, our algorithm shows very "stable" classification performance. Where "stable" means our algorithm can have classification performance close to the best result acheived by tuning and trying different numbers of demonstrations, which is random selection. This evaluation results align with our goal as we want to avoid the user having to guess and tune the number of demonstrations, and at the same time to let the user have a reasonable experience w.r.t the classification performance.

We also observe that different LLMs show different classification performance. From Table~\ref{tab:own_performance}, we observe that the Llama model shows better classification performance in general while the Mistral performance is low. For the Qwen model, the performance between random selection and our algorithm is close. It seems the Qwen model is able to have reasonable classification performance other than Zero-shot which is impressive. 

\paragraph{Performance gap to the best result in random selection} We compare the classification performance of our proposed algorithm with the best performance in random selection results. In our experiment setting, the random selection algorithm is more like a grid search. Since our proposed algorithm does not show dominant performance, we want to see how far our algorithm is from the best performance in random selection. The results are shown in Table~\ref{tab:comparison}. We check the gap model by model, and present the number demonstrations selected by our algorithm and the best number of demonstrations in random selection. We also provide the average classification performance of our proposed algorithm and the best performance in random selection for reference.

\begin{table}[h]
    \centering
    \begin{tabularx}{\textwidth}{l|c|c|c|c|c|c}
        \toprule
        \textbf{Dataset} & \textbf{Llama} & \textbf{Mistral} & \textbf{Qwen} & \textbf{Llama(R)} & \textbf{Mistral(R)} & \textbf{Qwen(R)} \\
        \hline
        tae & \textbf{0.28$\pm$0.1} & 0.21$\pm$0.1 & \underline{\textbf{0.28$\pm$0.07}} & 0.27$\pm$0.13 & 0.21$\pm$0.11 & 0.23$\pm$0.11 \\
        cmc & \underline{\textbf{0.12$\pm$0.03}} & 0.0$\pm$0.0 & 0.05$\pm$0.01 & 0.01$\pm$0.06 & 0.0$\pm$0.0 & 0.05$\pm$0.05\\
        wine & \underline{\textbf{0.37$\pm$0.04}} & 0.0$\pm$0.0 & \textbf{0.06$\pm$0.03} & 0.19$\pm$0.1 & 0.0$\pm$0.0 & 0.05$\pm$0.05 \\
        iris & 0.08$\pm$0.07 & 0.0$\pm$0.0 & 0.06$\pm$0.01 & \textbf{0.09$\pm$0.05} & 0.0$\pm$0.0 & \underline{\textbf{0.21$\pm$0.11}} \\
        penguins & 0.16$\pm$0.04 & 0.0$\pm$0.0 & 0.03$\pm$0.02 & \underline{\textbf{0.18$\pm$0.1}} & \textbf{0.01$\pm$0.02} & \textbf{0.14$\pm$0.07} \\
        LED & 0.06$\pm$0.03 & 0.0$\pm$0.0 & 0.05$\pm$0.05 & \textbf{0.08$\pm$0.04} & 0.0$\pm$0.0 & \underline{\textbf{0.12$\pm$0.07}}\\
        Customers & 0.26$\pm$0.04 & \textbf{0.09$\pm$0.08} & 0.14$\pm$0.09 & \underline{\textbf{0.49$\pm$0.33}} & 0.02$\pm$0.03 & \textbf{0.15$\pm$0.08} \\
        spambase & \underline{\textbf{0.29$\pm$0.04}} & \textbf{0.29$\pm$0.07} & \textbf{0.25$\pm$0.06} & 0.19$\pm$0.1 & 0.25$\pm$0.12 & 0.23$\pm$0.07\\
        \bottomrule
    \end{tabularx}
    \caption{Performance table. The bold values means better performance by comparing with same LLM but different demonstration selection strategies. The value with underline means the best performance for each dataset.}
    \label{tab:own_performance}
\end{table}

\begin{table}[h]
    \centering
    \begin{tabularx}{\textwidth}{l|c c | c c|c c | c c|c c | c c}
        \toprule
        \textbf{Dataset}   & \multicolumn{4}{c|}{\textbf{Llama}} & \multicolumn{4}{c|}{\textbf{Mistral}} & \multicolumn{4}{c}{\textbf{Qwen}} \\
        & \#d & avg & \#d & best   & \#d & avg & \#d & best & \#d & avg & \#d & best\\
        \hline
        tae        & 1  & 0.28 & 4  & 0.36 & 4  & 0.21 & 8  & 0.28 & 4    & 0.28 & 10  & 0.3  \\
        cmc        & 45 & 0.12 & 2  & 0.19 & 42 & 0    & 0  & 0     & 42   & 0.05 & 2   & 0.15 \\
        wine       & 1  & 0.37 & 6  & 0.26 & 5  & 0    & 0  & 0     & 4    & 0.06 & 2   & 0.15 \\
        iris       & 3  & 0.08 & 4  & 0.14 & 3  & 0    & 0  & 0     & 3    & 0.06 & 2   & 0.37 \\
        penguins   & 5  & 0.16 & 10 & 0.29 & 5  & 0    & 6  & 0.04  & 5   & 0.03 & 4   & 0.19 \\
        LED        & 5  & 0.06 & 6  & 0.11 & 4  & 0    & 0  & 0     & 4    & 0.05 & 10  & 0.19 \\
        Customers  & 1  & 0.26 & 10 & 0.86 & 3  & 0.09 & 4  & 0.06 & 3   & 0.14 & 4   & 0.21 \\
        spambase   & 3  & 0.29 & 8  & 0.29 & 1  & 0.29 & 10 & 0.34 & 1   & 0.25 & 2   & 0.21 \\
        \hline
    \end{tabularx}
    \caption{Classification performance gap between our proposed algorithm comparing to the best result from the random selection strategy with demonstration number less or equal than ten. The "Our" means the number of demonstration selected by our proposed algorithm. The "Perf" means the average Macro-F1 score. And the "Best" means the number of demonstrations which shows the best classification performance from the number set [0, 2, 4, 6, 8, 10].  }
    \label{tab:comparison}
\end{table}

\subsection{Spectral Gap}
\begin{figure}[h!]
    \centering
    \begin{subfigure}[b]{0.24\textwidth}
        \includegraphics[width=\textwidth]{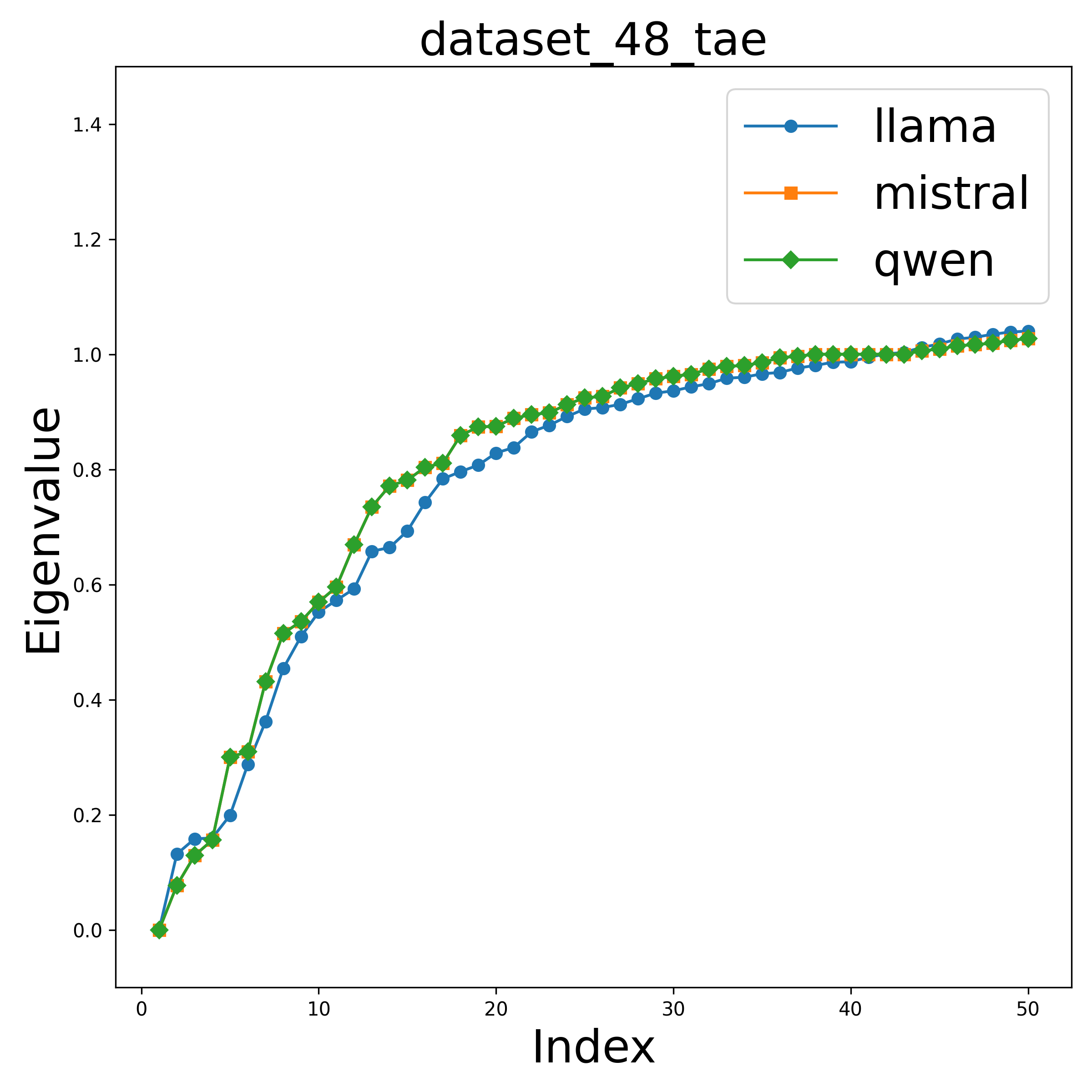}
    \end{subfigure}
    \begin{subfigure}[b]{0.24\textwidth}
        \includegraphics[width=\textwidth]{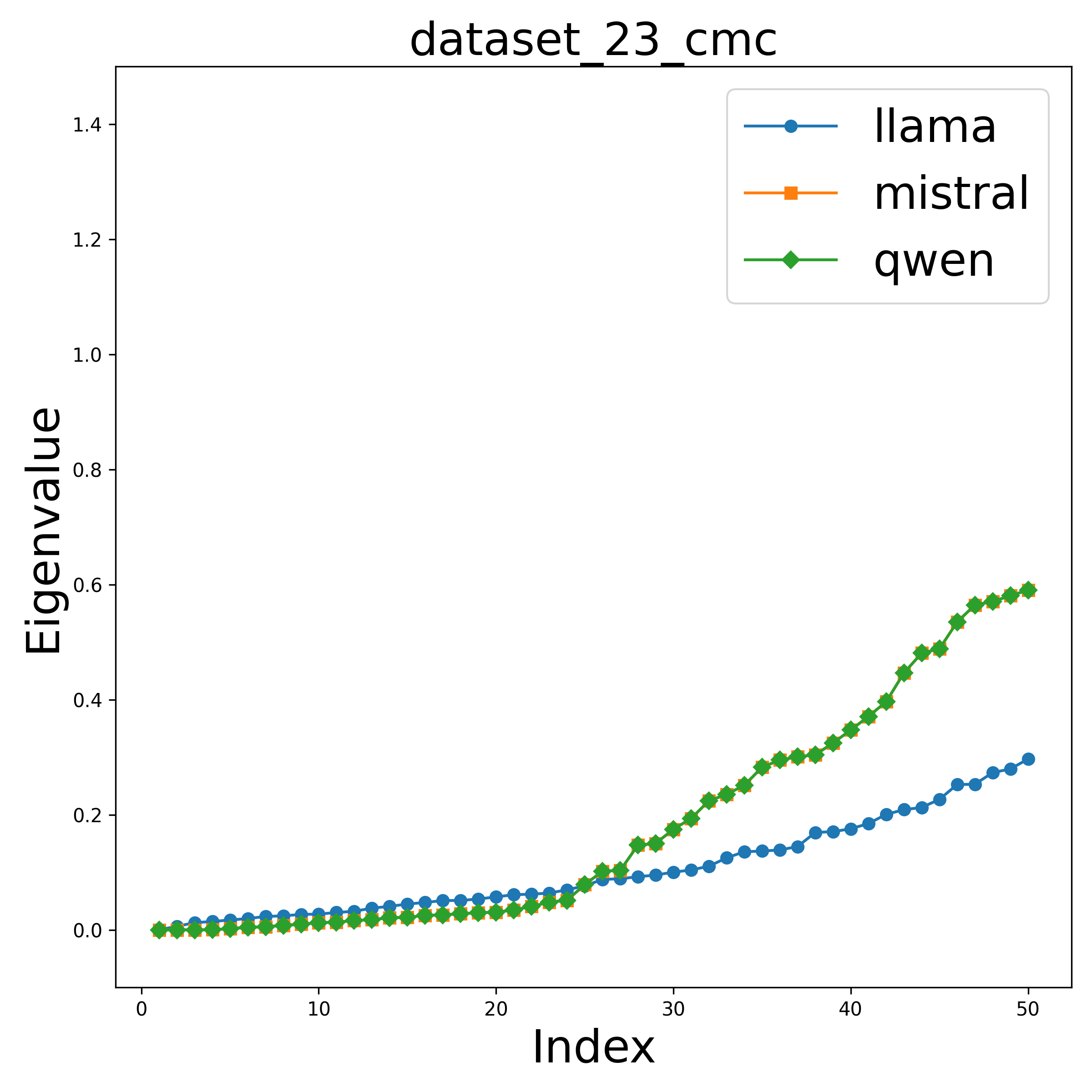}
    \end{subfigure}
        \begin{subfigure}[b]{0.24\textwidth}
        \includegraphics[width=\textwidth]{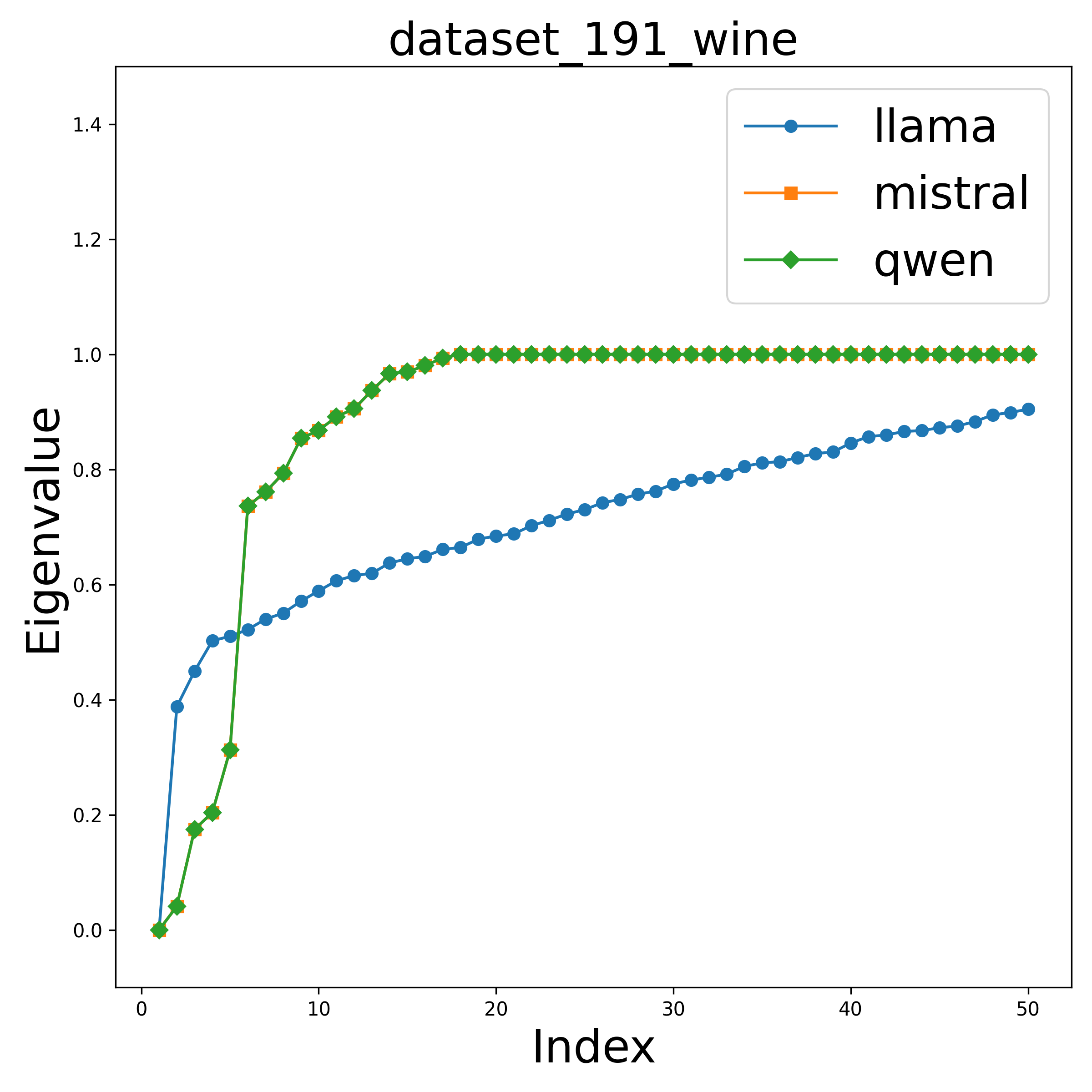}
    \end{subfigure}
        \begin{subfigure}[b]{0.24\textwidth}
        \includegraphics[width=\textwidth]{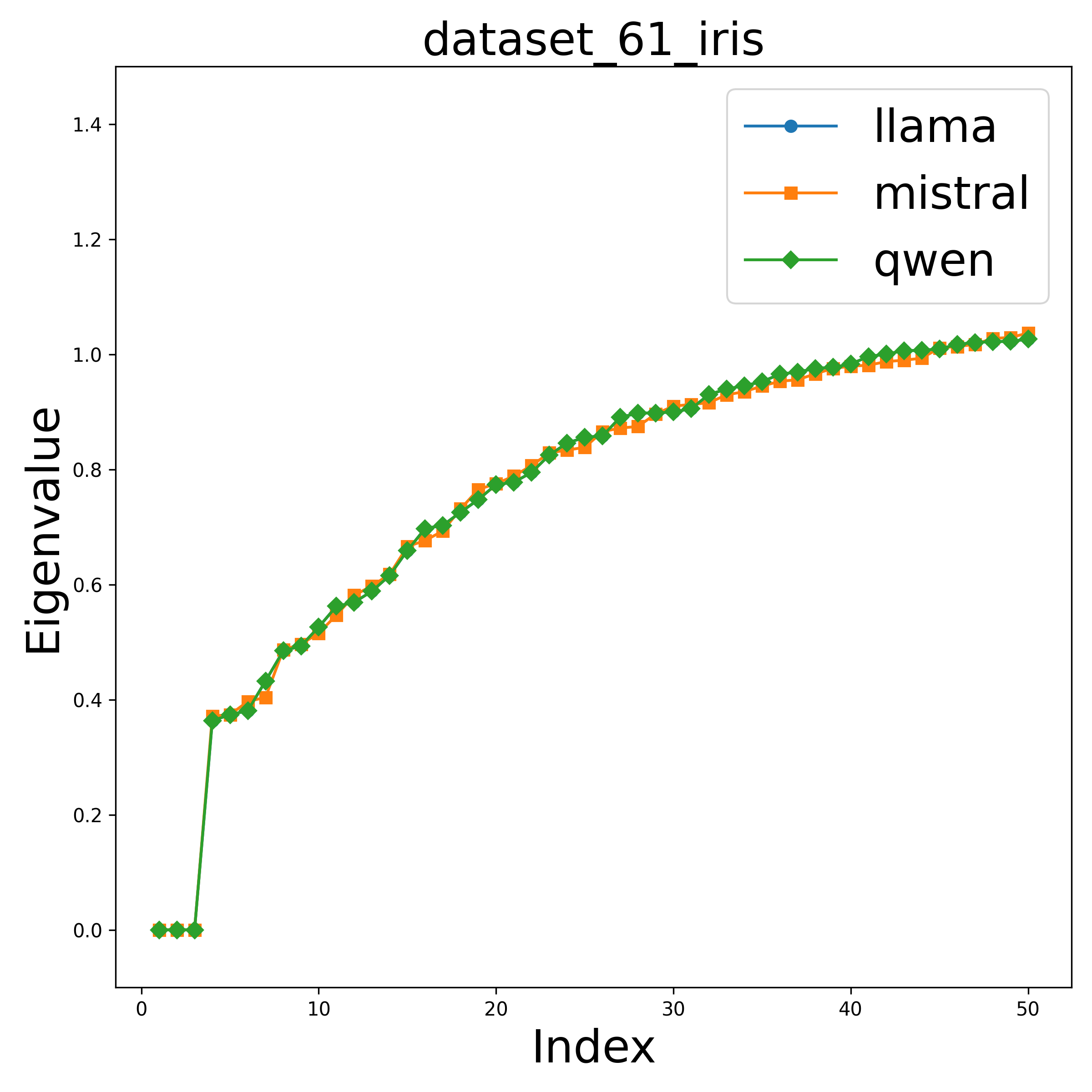}
    \end{subfigure}
    
    \vspace{0.1in} 
    
    \begin{subfigure}[b]{0.24\textwidth}
        \includegraphics[width=\textwidth]{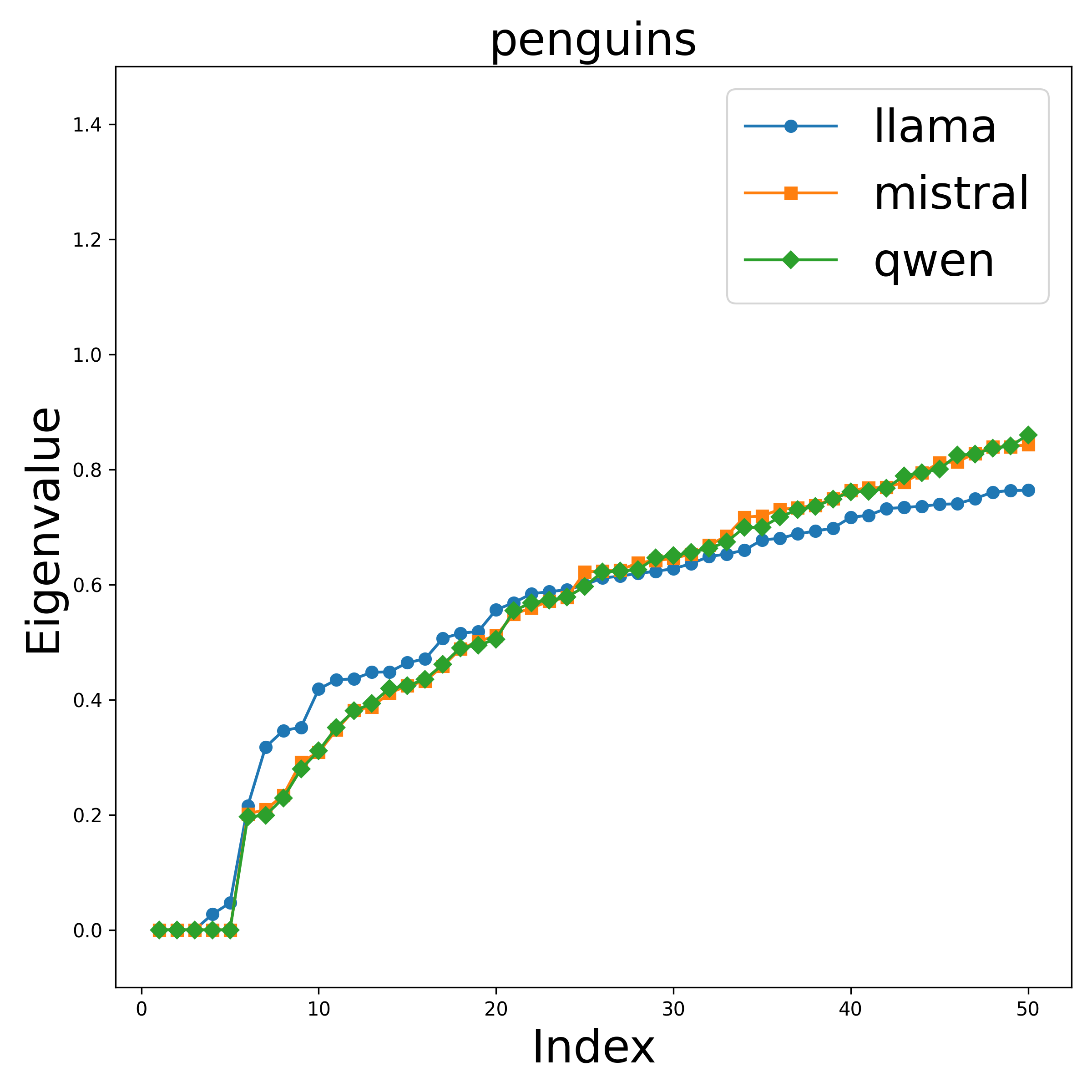}
    \end{subfigure}
    \hfill
    \begin{subfigure}[b]{0.24\textwidth}
        \includegraphics[width=\textwidth]{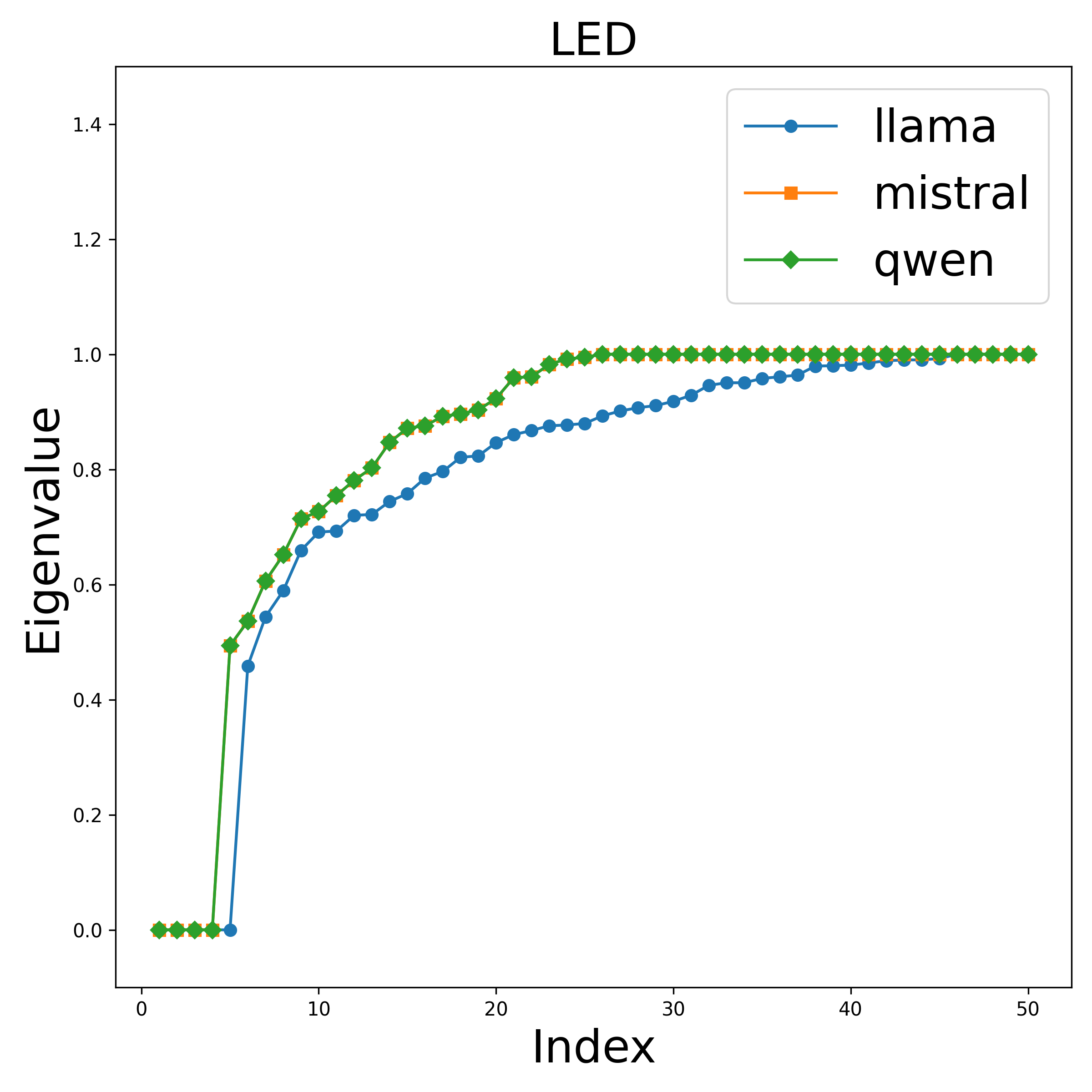}
    \end{subfigure}
    \hfill
        \begin{subfigure}[b]{0.24\textwidth}
        \includegraphics[width=\textwidth]{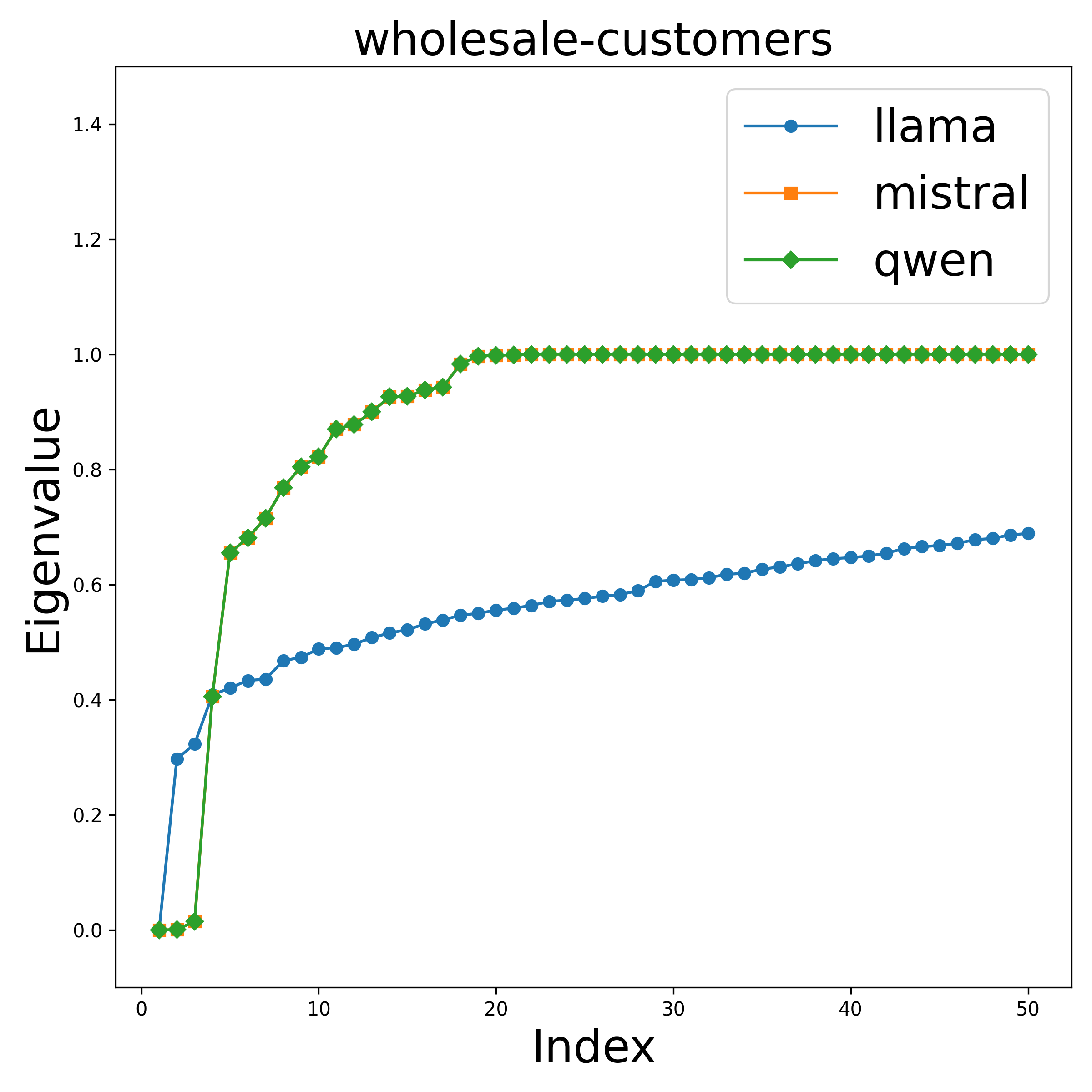}
    \end{subfigure}
    \hfill
        \begin{subfigure}[b]{0.24\textwidth}
        \includegraphics[width=\textwidth]{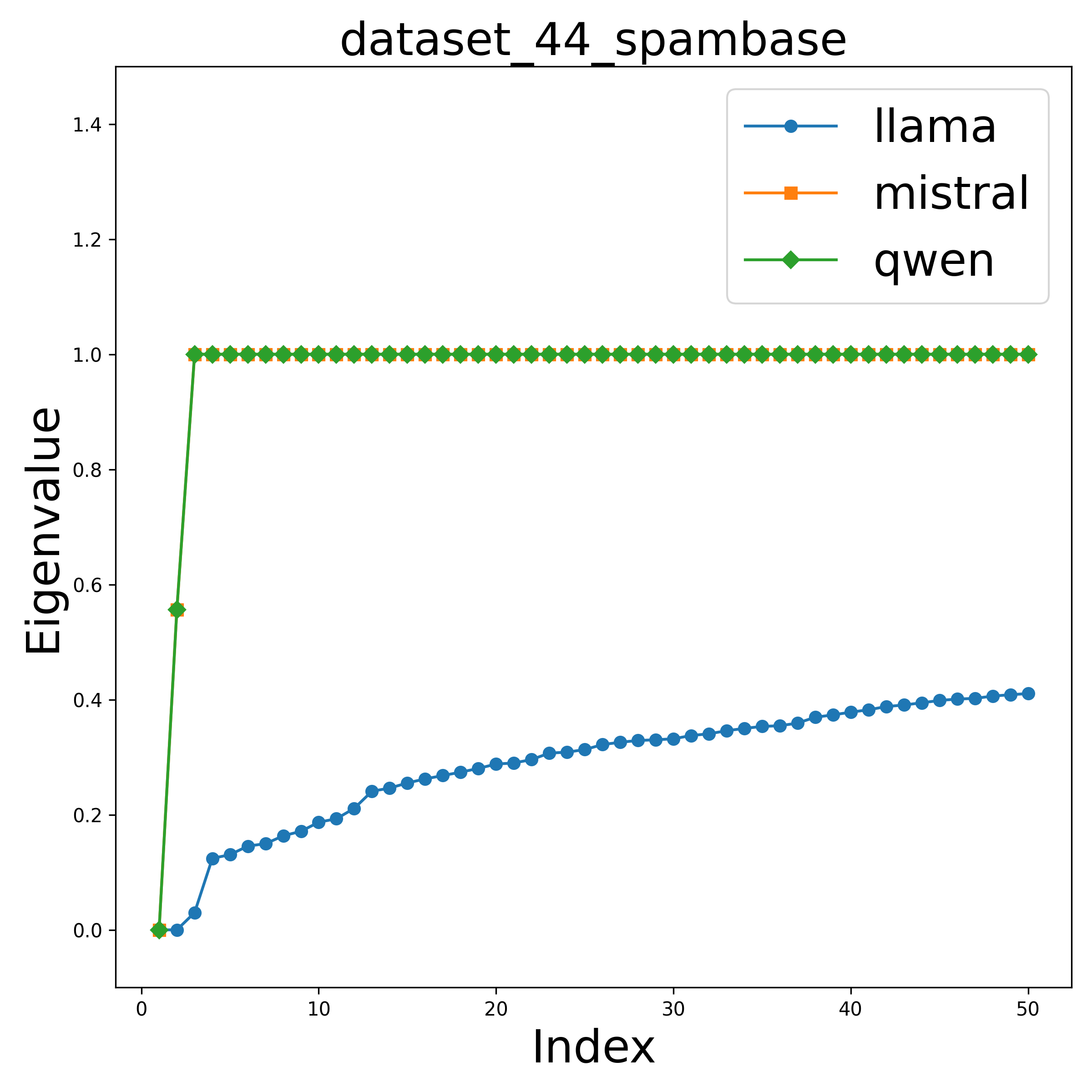}
    \end{subfigure}
    \caption{First 50 eigenvalues (sorted) of experiment datasets.}
    \label{fig:spectral_gap}
\end{figure}

To calculate the spectral gap, we use the code from Scikit-learn~\cite{pedregosa2011scikit} to calculate the eigenvalues and use the "arpack"~\cite{lehoucq1998arpack} numerical solver. We visualized the first 50 eigenvalues of each dataset in Figure~\ref{fig:spectral_gap}. As we can read from the figures, most datasets have obvious spectral gap except the \textit{tae} and \textit{cmc} datasets. 

\section{Related Work}
\paragraph{Classification tasks on tabular data with LLM} The Language-Interfaced Fine-Tuning(LIFT) work~\cite{dinh2022lift} introduces the study of applying LLM to non-language classification and regression tasks. It transfers each row of the tabular data into a plain English sentence with a fixed template, and fine-tunes the pre-trained LLM with them. With that, the LLM can achieve significant performance on classification and regression tasks. The work~\cite{hegselmann2023tabllm} focus on the classification task specifically and compared the performance with tree based algorithm. Unlike the previous works that treat every cell value as a string type, ~\cite{yan2024making} goes deeper and proposed a method to tokenize the numerical cell values into high dimensional vectors and fine tunes the pre-trained LLM to obtain better prediction performance. The work~\cite{nam2023semi} presents a semi-supervised method for ICL over tabular data without fine tune the pre-trained LLM and focus on the prompt engineering.

\paragraph{Demonstration Selection for In-Context Learning} The work~\cite{min2022rethinking} discusses the role of demonstrations and the importance of data labels in ICL. How to select demonstrations, or say samples, has been mentioned in few related works~\cite{qin2023context}~\cite{hu2024strategic}~\cite{zebaze2024context}~\cite{pmlr-v139-zhao21c}~\cite{liu-etal-2022-makes}~\cite{rubin-etal-2022-learning}~\cite{wang2023large}. For example, in the Section 3.2 of ~\cite{dinh2022lift}, the authors mentioned that their algorithm needs more demonstrations while the number of classes increasing. The work~\cite{wang2023large} introduces a Bayesian approach to select demonstrations for ICL. The work~\cite{hu2024strategic} presents a demonstration selection method for ICL with a fairness concern.

\section{Conclusion}
In this work, we present an algorithm to select the number of demonstrations automatically for the ICL based classification task of tabular data. The core idea of our algorithm is to estimate the data distribution of the input date in the LLM's representation space. Here we use the token IDs of the LLM. We estimate the number of clusters by using the spectral gap method from the spectral graph theory, and use it as the number of selected demonstrations. Moreover, we design a prompt template for the ICL based classification task. 

Our proposed algorithm has low computational cost since it only needs the token ids of the demonstrations instead of the embedding vectors. With our proposed algorithm, we can avoid the guess and check process when selecting the number of demonstrations. We also show that our algorithm can achieve reasonable classification performance on eight public available tabular datasets with three different LLMs. While our method does not exhibit superior performance relative to random selection strategies, it consistently yields stable results that approximate the optimal performance achieved through random selection.

In the future work, we plan to extend our algorithm to large scale tabular data with possible missing values, and to explore the capabilities of much larger LLMs.



\medskip

\bibliographystyle{agsm}
\bibliography{references}

\end{document}